\documentclass[runningheads]{llncs}
\usepackage{graphicx}
\usepackage{subcaption}
\usepackage{graphicx}
\usepackage{hyperref}
\usepackage{amsmath}
\usepackage{booktabs}
\usepackage{multicol}
\usepackage{multirow}
\usepackage{tabularx}
\usepackage{tabu}
\usepackage{lipsum}
\usepackage{array}

\newcolumntype{M}[1]{>{\centering\arraybackslash}m{#1}}
\newcolumntype{P}[1]{>{\centering\arraybackslash}p{#1}}

\usepackage[table]{xcolor}
\usepackage{colortbl}
\usepackage{xcolor}

\begin{document}
\title{Guiding the Guidance: A Comparative Analysis of User Guidance 
Signals for Interactive
Segmentation of Volumetric Images}
\titlerunning{Guiding the Guidance: A Comparative Analysis of User Guidance Signals}

\author{Anonymous}
\institute{Anonymous Organization \\
\email{**@******.***}}
%
\author{Zdravko Marinov\inst{1,2} \and
Rainer Stiefelhagen\inst{1} \and
Jens Kleesiek\inst{3,4}}
\authorrunning{Z. Marinov et al.}
%
\institute{Karlsruhe Institute of Technology, Karlsruhe, Germany \\
\and
HIDSS4Health - Helmholtz Information and Data Science School for Health,
Karlsruhe/Heidelberg, Germany \\ 
\and 
Institute for AI in Medicine, University Hospital Essen, Essen, Germany \\ 
\and
Cancer Research Center Cologne Essen (CCCE), University Medicine Essen, Essen,\\
\email{$^1$\{firstname.lastname@kit.edu\} , $^3$\{firstname.lastname@uk-essen.de\}}
}
\maketitle              
\begin{abstract}

Interactive segmentation reduces the annotation time of medical images and allows annotators to iteratively refine labels with corrective interactions, such as clicks. While existing interactive models transform clicks into user guidance signals, which are combined with images to form (image, guidance) pairs, the question of how to best represent the guidance has not been fully explored. To address this, we conduct a comparative study of existing guidance signals by training interactive models with different signals and parameter settings to identify crucial parameters for the model's design. Based on our findings, we design a guidance signal that retains the benefits of other signals while addressing their limitations. We propose an adaptive Gaussian heatmap guidance signal that utilizes the geodesic distance transform to dynamically adapt the radius of each heatmap when encoding clicks. We conduct our study on the MSD Spleen and the AutoPET datasets to explore the segmentation of both anatomy (spleen) and pathology (tumor lesions). Our results show that choosing the guidance signal is crucial for interactive segmentation as we improve the performance by 14\% Dice with our adaptive heatmaps on the challenging AutoPET dataset when compared to non-interactive models. This brings interactive models one step closer to deployment in clinical workflows. We will make our code publicly available.

\end{abstract}


\section{Introduction}

Deep learning models have achieved remarkable success in segmenting anatomy and lesions from medical images but often rely on large-scale manually annotated datasets~\cite{autopet}, \cite{MSD-Spleen}, \cite{BRATS}. This is challenging when working with volumetric medical data as voxelwise labeling requires a lot of time and expertise. Interactive segmentation models address this issue by utilizing weak labels, such as clicks, instead of voxelwise annotations~\cite{MIDeepSeg}, \cite{i3deep}, \cite{ECONet}. The clicks are transformed into guidance signals, e.g., Gaussian heatmaps or Euclidean/Geodesic distance maps, and used together with the image as a joint input for the interactive model. Annotators can make additional clicks in missegmented areas to iteratively refine the segmentation mask, which often significantly improves the prediction compared to non-interactive models~\cite{virtual-raters}, \cite{ITIS}. However, prior research on choosing guidance signals for interactive models is limited to small ablation studies~\cite{MIDeepSeg}, \cite{cvpr-study}, \cite{RITM}. There is also no systematic framework for comparing guidance signals, which includes not only accuracy but also efficiency and the ability to iteratively improve predictions with new clicks, which are all important aspects of interactive models~\cite{ECONet}. We address these challenges with the following contributions:

\begin{enumerate}
    \item We compare 5 existing guidance signals on the AutoPET~\cite{autopet} and MSD Spleen~\cite{MSD-Spleen} datasets and vary various hyperparameters. We show which parameters are essential to tune for each guidance and suggest default values.
    \item We introduce 5 guidance evaluation metrics \textbf{(M1)-(M5)}, which evaluate the performance, efficiency, and ability to improve with new clicks. This provides a systematic framework for comparing guidance signals in future research.
    \item Based on our insights from 1., we propose novel adaptive Gaussian heatmaps, which use geodesic distance values around each click to set the radius of each heatmap. Our adaptive heatmaps mitigate the weaknesses of the 5 guidances and achieve the best performance on AutoPET~\cite{autopet} and MSD Spleen~\cite{MSD-Spleen}.
\end{enumerate}

\textbf{Related Work.} Previous work comparing guidance signals has mostly been limited to small ablation studies. Sofiiuk et al.~\cite{RITM} and Benenson et al.~\cite{cvpr-study} both compare Euclidean distance maps with solid disks and find that disks perform better. However, neither of them explore different parameter settings for each guidance and both work with natural 2D images. Dupont et al.~\cite{UCP-Net} note that a comprehensive comparison of existing guidance signals would be helpful in designing interactive models. The closest work to ours is MIDeepSeg~\cite{MIDeepSeg}, which propose an user guidance based on exponentialized geodesic distances and compare it to existing guidance signals. However, they use only initial clicks and do not add iterative corrective clicks to refine the segmentation. In contrast to previous work, our research evaluates the influence of hyperparameters for guidance signals and assesses the guidances' efficiency and ability to improve with new clicks, in addition to accuracy. While some previous works~\cite{FCA} \cite{dynamic-embc}, \cite{dynamic-icip} propose using a larger radius for the first click's heatmap, our adaptive heatmaps offer a greater flexibility by adjusting the radius at each new click dynamically.


\section{Methods}
\subsection{Guidance Signals}
\label{sec:guidance_signals}


We define the five guidance signals over a set of $N$ clicks $\mathcal{C}=\{c_1,c_2,...,c_N\}$ where $c_i = (x_i,y_i,z_i)$ is the $i^{th}$ click. As disks and heatmaps can be computed independently for each click, they are defined for a single click $c_i$ over 3D voxels $v = (x,y,z)$ in the volume. The \textbf{disk} signal fills spheres with a radius $\sigma$ centered around each click $c_i$, which is represented by the equation in Eq. (\ref{eq:disks}).

\begin{equation}
  \text{disk}(v, c_i, \sigma) = \left \{
  \begin{aligned}
    & 1, && \text{if}\  ||v - c_i||_2 \leq \sigma \\
    & 0, && \text{otherwise}
  \end{aligned} \right.
  \label{eq:disks}
\end{equation} 

The \textbf{Gaussian heatmap} applies Gaussian filters centered around each click to create softer edges with an exponential decrease away from the click (Eq. (\ref{eq:heatmaps})).
\begin{equation}
  \text{heatmap}(v, c_i, \sigma) = \exp(-\frac{||v - c_i||_2}{2\sigma^2})
  \label{eq:heatmaps}
\end{equation} 

The \textbf{Euclidean distance transform} (EDT) is defined in Eq. (\ref{eq:edt}) as the minimum Euclidean distance between a voxel $v$ and the set of clicks $\mathcal{C}$. It is similar to the disk signal in Eq. (\ref{eq:disks}), but instead of filling the sphere with a constant value it computes the distance of each voxel to the closest click point. 

\begin{equation}
    \text{EDT}(v, \mathcal{C}) = \min_{c_i \in \mathcal{C}} ||v -c_i||_2
    \label{eq:edt}
\end{equation}

The \textbf{Geodesic distance transform} (GDT) is defined in Eq. (\ref{eq:gdt}) as the shortest path distance between each voxel in the volume and the closest click in the set $\mathcal{C}$~\cite{GeoS}. The shortest path in GDT also takes into account intensity differences between voxels along the path. We use the method of Asad et al.~\cite{FastGeodis} to compute the shortest path which is denoted as $\Phi$ in Eq. (\ref{eq:gdt}).

\begin{equation}
    \text{GDT}(v, \mathcal{C}) = \min_{c_i \in \mathcal{C}} \Phi(v, c_i)
    \label{eq:gdt}
\end{equation}

We also examine the \textbf{exponentialized Geodesic distance} (exp-GDT) proposed in MIDeepSeg~\cite{MIDeepSeg} that is defined in Eq. (\ref{eq:exp_gdt}) as an exponentiation of GDT: 

\begin{equation}
    \text{exp-GDT}(v, \mathcal{C}) = 1 - \exp(-\text{GDT}(v, \mathcal{C}))
    \label{eq:exp_gdt}
\end{equation}

\textit{Note:} We normalize signals to $[0,1]$ and invert intensity values for Euclidean and Geodesic distances $d(x)$ by $1 - d(x)$ for better highlighting of small distances.

We define our \textbf{adaptive Gaussian heatmaps} $\text{ad-heatmap}(v, c_i, \sigma_i)$ via:
\begin{equation}
\sigma_i=\lfloor a e^{-bx} \rfloor, \text{where \    }x=\frac{1}{|\mathcal{N}_{c_i}|}\sum_{v \in \mathcal{N}_{c_i}} \text{GDT}(v, \mathcal{C})
\label{eq:adaptive_heatmaps}
\end{equation}

Here, $\mathcal{N}_{c_i}$ is the 9-neighborhood of $c_i$, $a=13$ limits the maximum radius to 13, and $b=0.15$ is set empirically\footnote{We note that $b$ can also be automatically learned but we leave this for future work.} (details in supplementary). The radius $\sigma_i$ is smaller for higher $x$, i.e., when the mean geodesic distance in the neighboring voxels is high, indicating large intensity changes such as edges. This leads to a preciser guidance with a smaller radius $\sigma_i$ near edges and a larger radius in homogeneous areas such as clicks in the center of the object of interest. An example of this process and each guidance signal can be seen in Figure \ref{fig:results}a).

\subsection{Model Backbone and Datasets}
\label{sec:model_and_datasets}
We use the DeepEdit~\cite{DeepEdit} model with a U-Net backbone~\cite{UNet} and simulate a fixed number of clicks $N$ during training and evaluation. For each volume, $N$ clicks are iteratively sampled from over- and undersegmented predictions of the model as in \cite{Sakinis} and represented as foreground and background guidance signals. We implemented our experiments with MONAI Label~\cite{MONAI-Label} and will release our code.

We trained and evaluated all of our models on the openly available AutoPET~\cite{autopet} and MSD Spleen~\cite{MSD-Spleen} datasets. MSD Spleen~\cite{MSD-Spleen} contains 41 CT volumes with voxel size $0.79\times0.79\times5.00$mm$^3$ and average resolution of $512\times512\times89$ voxels with dense annotations of the spleen. AutoPET~\cite{autopet} consists of 1014 PET/CT volumes with annotated tumor lesions of melanoma, lung cancer, or lymphoma. We discard the 513 tumor-free patients, leaving us with 501 volumes. We also only use PET data for our experiments. The PET volumes have a voxel size of $2.0\times2.0\times2.0$mm$^3$ and an average resolution of $400\times400\times352$ voxels. 

\subsection{Hyperparameters: Experiments}
\label{sec:hyperparameters}
We keep these parameters constant for all models: $\text{learning rate}=10^{-5}$, \#clicks $N=10$,  Dice Cross-Entropy Loss~\cite{nnUNet}, and a fixed 80-20 training-validation split ($\mathcal{D}_{\text{train}} / \mathcal{D}_{\text{val}})$. We apply the same data augmentation transforms to all models and simulate clicks as proposed in Sakinis et al.~\cite{Sakinis}. We train using one A100 GPU for 20 and 100 epochs on AutoPET~\cite{autopet} and MSD Spleen~\cite{MSD-Spleen} respectively.

We vary the following four hyperparameters \textbf{(H1) -- (H4)}:

\textbf{(H1) Sigma.} We vary the radius $\sigma$ of disks and heatmaps in Eq. (\ref{eq:disks}) and (\ref{eq:heatmaps}) and also explore how this parameter influences the performance of the distance-based signals in Eq. (\ref{eq:edt})-(\ref{eq:exp_gdt}). Instead of initializing the seed clicks $\mathcal{C}$ as individual voxels $c_i$, we initialize the set of seed clicks $\mathcal{C}$ as all voxels within a radius $\sigma$ centered at each $c_i$ and then compute the distance transform as in Eq. (\ref{eq:edt})--(\ref{eq:exp_gdt}). 

\textbf{(H2) Theta.} We explore how truncating the values of distance-based signals in Eq. (\ref{eq:edt})--(\ref{eq:exp_gdt}) affects the performance. We discard the top $\theta \in \{10\%, 30\%, 50\%\}$ of the distance values and keep only smaller distances closer to the clicks making the guidance more precise. Unlike MIDeepSeg~\cite{MIDeepSeg}, we compute the $\theta$ threshold for each image individually, as fixed thresholds may not be suitable for all images.

\textbf{(H3) Input Adaptor.} We test three methods for combining guidance signals with input volumes proposed by Sofiuuk et al. \cite{RITM} - Concat, Distance Maps Fusion (DMF), and Conv1S. Concat combines input and guidance by concatenating their channels. DMF additionally includes $1\times1$ conv. layers to adjust the channels to match the original size in the backbone. Conv1S has two branches for the guidance and volume, which are summed and fed to the backbone.

\textbf{(H4) Probability of Interaction.} We randomly decide for each volume whether to add the $N$ clicks or not, with a probability of $p$, in order to make the model more independent of interactions and improve its initial segmentation. All the hyperparameters we vary are summarized in Table \ref{tab:hyperparameters}. Each combination of hyperparameters corresponds to a separately trained DeepEdit~\cite{DeepEdit} model.

\begin{table}[h]
    \centering
    \vspace*{-0.7cm}
    \caption{Variation of hyperparameters \textbf{(H1) -- (H4)} in our experiments.}
    \scalebox{1.0}{
    \begin{tabular}{cc|cc}
    \toprule
        $\sigma$ & $\{0, 1, 5, 9, 13\}$ \ \ &
       \ \ \ \ \ \ $\theta$ \ \ \ \ \ \ & $\{0\%, 10\%, 30\%, 50\%\}$ \\ 
        Input Adaptor & $\{$Concat, DMF, Conv1S$\}$ \ \ &
        $p$ & $\{50\%, 75\%, 100\%\}$ \\
    \bottomrule 
    \end{tabular}}
    \label{tab:hyperparameters}
\end{table}

\subsection{Additional Evaluation Metrics}
\label{sec:guidance_profiles}
We use 5 metrics \textbf{(M1)--(M5)} (Table \ref{tab:metrics}) to evaluate the validation performance.

\vspace{-0.6cm}
\begin{table}[h!]
\rowcolors{1}{gray!0}{gray!15}
\centering
\caption{Evaluation metrics for the comparison of guidance signals.}
\scalebox{0.85}{
\renewcommand*{\arraystretch}{1.3}

\begin{tabular}
{m{0.1\linewidth}m{0.16\linewidth}|m{0.85\linewidth}}
\toprule
 \multicolumn{2}{c|}{Metric} & \multicolumn{1}{c}{Description} \\
\hline
\centering \textbf{(M1)} & \centering Final Dice & Mean Dice score after $N=10$ clicks per volume. \\
\centering \textbf{(M2)} & \centering Initial Dice & Mean Dice score before any clicks per volume $(N=0)$. A higher initial Dice indicates less work for the annotator. \\
\centering \textbf{(M3)} & \centering Efficiency & Inverted$^{*}$ time measurement $(1 - T)$ in seconds needed to compute the guidance. Low efficiency increases the annotation time with every new click. Note that this metric depends on the volume size and hardware setup. $^{*}$Our maximum measurement $T_{\text{max}}$ is shorter than 1 second. \\
\centering \textbf{(M4)} & \centering Consistent Improvement & Ratio of clicks $\mathcal{C}^{+}$ that improve the Dice score to the total number of validation clicks: $\frac{|\mathcal{C}^{+}|}{N\cdot|\mathcal{D}_{\text{val}}|}$, where $N=10$ and $\mathcal{D}_{\text{val}}$ is the validation dataset.  \\
\centering \textbf{(M5)} & \centering Ground-truth Overlap & Overlap of the guidance $G$ with the ground-truth mask $M$: $\frac{|M \cap G|}{|G|}$. This estimates the guidance precision as corrective clicks are often near boundaries, and if guidances are too large, such as disks with a large $\sigma$, there is a large overlap with the background outside the boundary.  \\
\bottomrule
\end{tabular}}
\vspace{-1.0cm}

\label{tab:metrics}
\end{table}

\section{Results}

\subsection{Hyperparemeters: Results}
\label{sec:varied_hyperparameters}
\begin{figure}[]
    \centering
    \includegraphics[width=\textwidth]{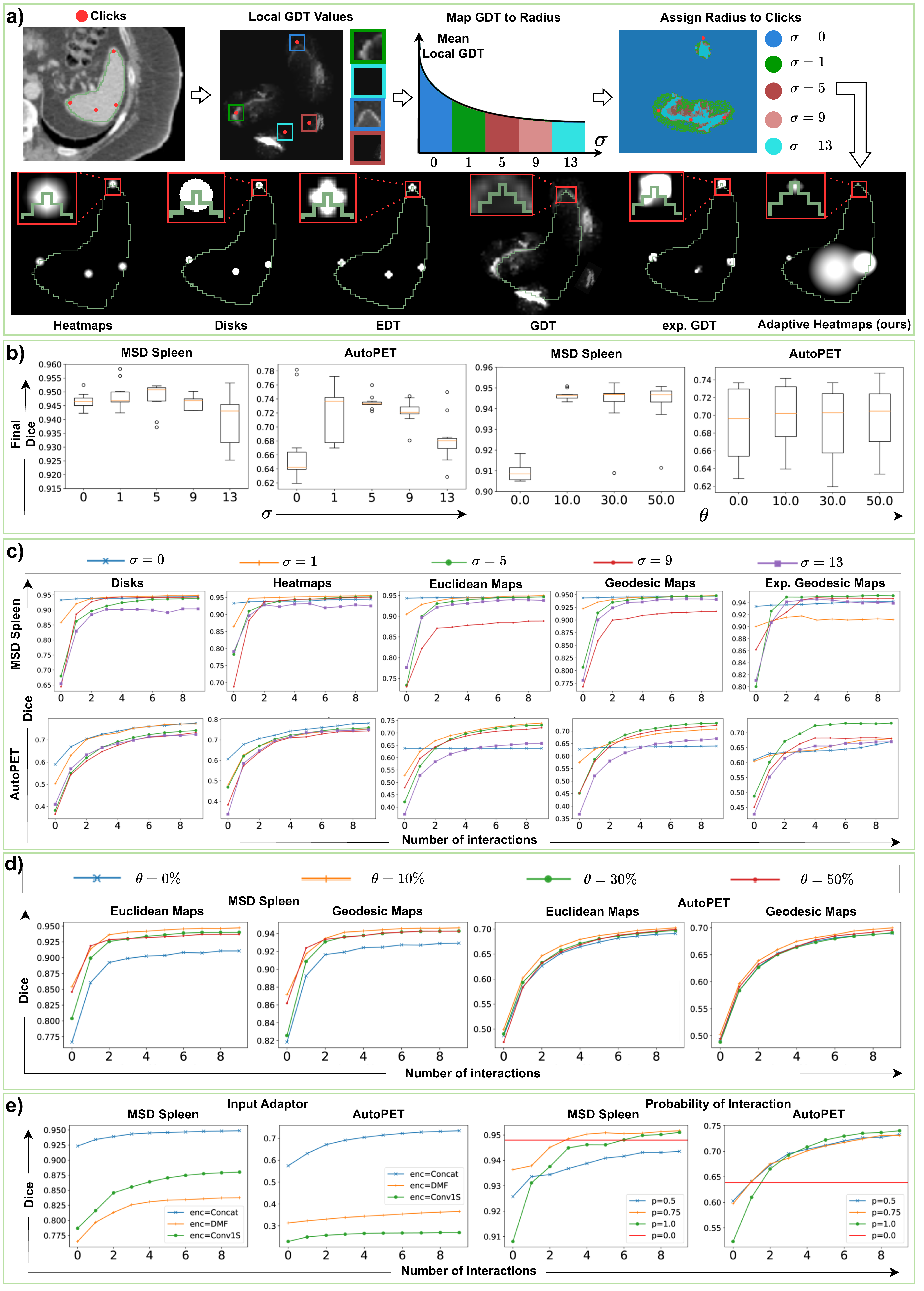}
    \caption{a) Our adaptive heatmaps use the mean local GDT values around each click to compute click-specific radiuses and form larger heatmaps in homogeneous regions and smaller near boundaries. b)--e) Results for varied hyperparameters. b) The final Dice of $\sigma$ (left) and $\theta$ (right) aggregated for all guidance signals. c) The influence of $\sigma$ and d) the influence of $\theta$ on individual guidances. e) Results for the different input adaptors (left) and probability of interaction $p$ (right).}
    \label{fig:results}
\end{figure}

We first train a DeepEdit~\cite{DeepEdit} model for each $(\sigma, \theta)$ pair and set $p=100\%$ and the input adaptor to Concat to constrain the parameter space.

\textbf{(H1) Sigma.} Results in Fig. \ref{fig:results}b) show that on MSD Spleen~\cite{MSD-Spleen}, the highest Dice scores are at $\sigma=5$, with a slight improvement for two samples at $\sigma=1$, but performance decreases for higher values $\sigma>5$. On AutoPET~\cite{autopet}, $\sigma=5$ and two samples with $\sigma=0$ show the best performance, while higher values again demonstrate a significant performance drop. Fig. \ref{fig:results}c) shows that the best final and initial Dice for disks and heatmaps are with $\sigma=1$ and $\sigma=0$. Geodesic maps exhibit lower Dice scores for small $\sigma < 5$ and achieve the best performance for $\sigma=5$ on both datasets. Larger $\sigma$ values lead to a worse initial Dice for all guidance signals. Differences in results for different $\sigma$ values are more pronounced in AutoPET~\cite{autopet} as it is a more challenging dataset~\cite{AutoPET-Submission-1}, \cite{AutoPET-Submission-2}, \cite{AutoPET-Winner}, \cite{isbi_2023}.

\textbf{(H2) Theta.} We examine the impact of truncating large distance values for the EDT and GDT guidances from Eq. (\ref{eq:edt}) and (\ref{eq:gdt}). Fig. \ref{fig:results}a) shows that the highest final Dice scores are achieved with $\theta=10$ for MSD Spleen~\cite{MSD-Spleen}. On AutoPET~\cite{autopet}, the scores are relatively similar when varying $\theta$ with a slight improvement at $\theta=10$. The results in Fig.~\ref{fig:results}d) also confirm that $\theta=10$ is the optimal parameter for both datasets and that not truncating values on MSD Spleen~\cite{MSD-Spleen}, i.e. $\theta=0$, leads to a sharp drop in performance.

\begin{table}[t]
    \centering
    \caption{Optimal parameter settings of our interactive models and the non-interactive baseline (non-int.) and their Dice scores. }
    \scalebox{0.8}{
    \begin{tabular}{c|c|cccccc|c|cccccc}
    \toprule
        {} & \multicolumn{7}{c|}{\scriptsize MSD Spleen~\cite{MSD-Spleen}} & \multicolumn{7}{c}{\scriptsize AutoPET~\cite{autopet}} \\ 
       
        {} & \scriptsize non-int. & \scriptsize Disks & \scriptsize  Heatmaps & \scriptsize  EDT & \scriptsize  GDT & \scriptsize  exp-GDT & \scriptsize Ours & \scriptsize non-int. & \scriptsize Disks & \scriptsize  Heatmaps & \scriptsize  EDT & \scriptsize  GDT & \scriptsize  exp-GDT & \scriptsize Ours  \\ \hline
       \scriptsize $\sigma$ & \scriptsize - &\scriptsize 1 & \scriptsize 1 & \scriptsize 1 & \scriptsize 5 & \scriptsize 5 & \scriptsize adaptive & \scriptsize - & \scriptsize 0 & \scriptsize 0 & \scriptsize 1 & \scriptsize 5 & \scriptsize 5 & \scriptsize adaptive \\
       \scriptsize $\theta$ & \scriptsize - &\scriptsize - & \scriptsize - & \scriptsize - & \scriptsize 10\% & \scriptsize 10\% & \scriptsize - & \scriptsize - &\scriptsize - & \scriptsize - & \scriptsize 10\% & \scriptsize 10\% & \scriptsize - & \scriptsize - \\
       \scriptsize Adaptor & \scriptsize Concat & \multicolumn{6}{c|}{\scriptsize Concat} & \scriptsize Concat & \multicolumn{6}{c}{\scriptsize Concat} \\ 
        \scriptsize $p$ & \scriptsize 0\% & \multicolumn{6}{c|}{\scriptsize 75\%} & \scriptsize 0\% & \multicolumn{6}{c}{\scriptsize 100\%} \\ 
       
       \scriptsize Dice & \scriptsize 94.90 & \scriptsize 95.91 & \scriptsize 95.82  & \scriptsize 95.81 & \scriptsize 95.19 & \scriptsize 95.22 & \textbf{\scriptsize 96.87} & \scriptsize 64.89 & \scriptsize 78.15 & \scriptsize 78.24 & \scriptsize 75.22 & \scriptsize 74.50 & \scriptsize 73.19  &  \textbf{\scriptsize 79.89} \\
    \bottomrule
    \end{tabular}}
    \label{tab:sigma_theta_results}
\end{table}

For our next experiments, we fix the optimal $(\sigma, \theta)$ pair for each of the five guidances (see Table \ref{tab:sigma_theta_results}) and train a DeepEdit~\cite{DeepEdit} model for all combinations of input adaptors and probability of interaction.

\textbf{(H3) Input Adaptor.} We look into different ways of combining guidance signals with input volumes using the input adaptors proposed by Sofiuuk et al.~\cite{RITM}. The results in Fig. \ref{fig:results}e) indicate that the best performance is achieved by simply concatenating the guidance signal with the input volume. This holds true for both datasets and the difference in performance is substantial.

\textbf{(H4) Probability of Interaction.} Fig. \ref{fig:results}e) shows that $p \in \{75\%, 100\%\}$ results in the best performance on MSD Spleen~\cite{MSD-Spleen}, with a faster convergence rate for $p=75\%$. However, with $p=50\%$, the performance is worse than the non-interactive baseline ($p=0\%$). On AutoPET~\cite{autopet}, the results for all $p$ values are similar, but the highest Dice is achieved with $p=100\%$. Note that $p=100\%$ results in lower initial Dice scores and requires more interactions to converge, indicating that the models depend more on the interactions. For the rest of our experiments, we use the optimal hyperparameters for each guidance in Table \ref{tab:hyperparameters}.

\subsection{Additional Evaluation Metrics: Results}
\label{sec:guidance_profiles_results}
\begin{figure}[t]
    \centering
    \includegraphics[width=0.85\textwidth]{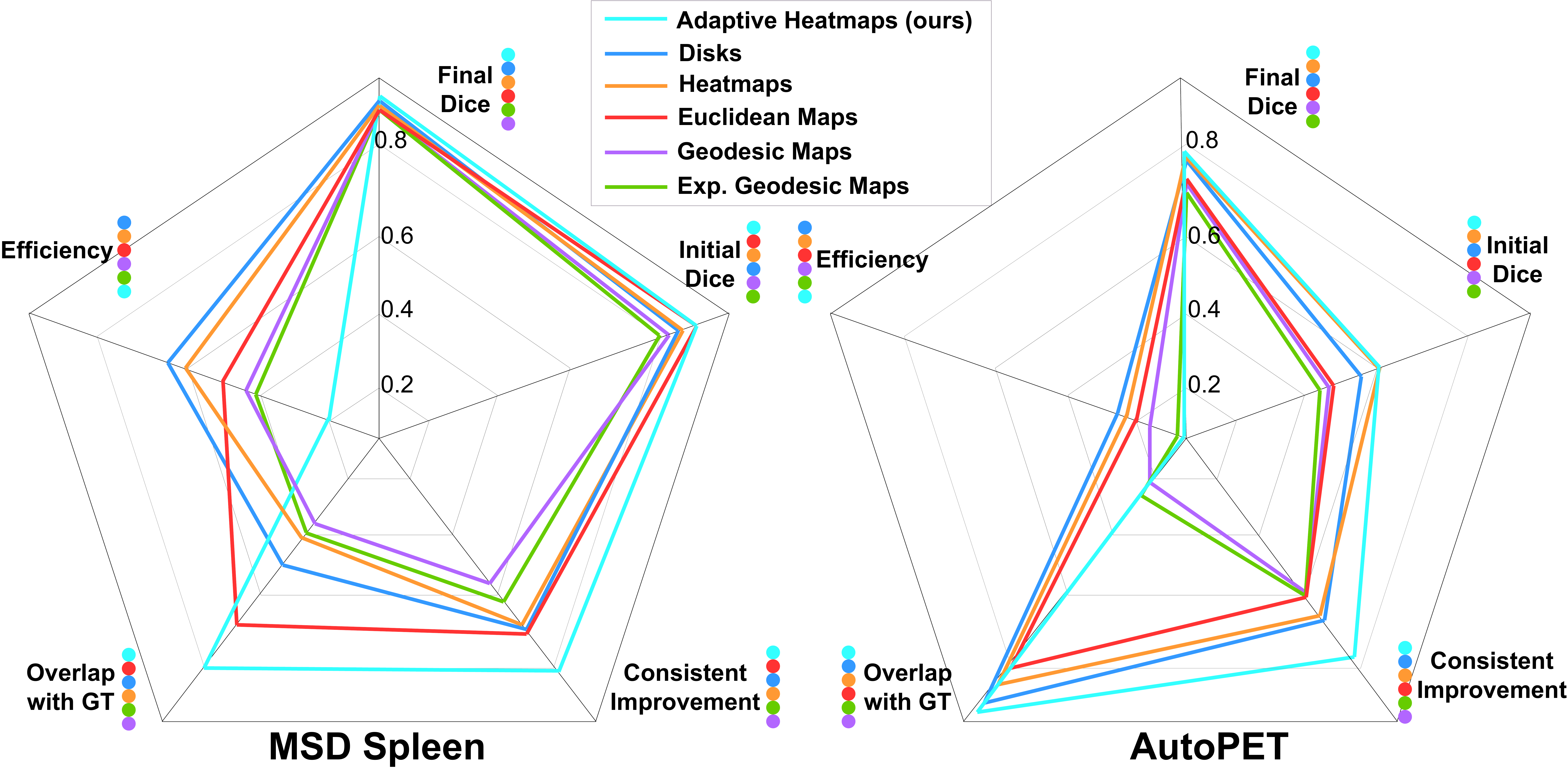}
    \caption{Comparison of all guidance signals with our 5 metrics. The circles next to each metric represent the ranking of the guidances (sorted top-to-bottom).}
    \label{fig:spider_plots}
\end{figure}

The comparison of the guidance signals using our five metrics \textbf{(M1)--(M5)} can be seen in Fig. \ref{fig:spider_plots}. Although the concrete values for MSD Spleen~\cite{MSD-Spleen} and AutoPET~\cite{autopet} are different, the five metrics follow the same trend on both datasets.

\textbf{(M1) Initial and (M2) Final Dice.} Overall, all guidance signals improve their initial-to-final Dice scores after $N$ clicks, with AutoPET~\cite{autopet} showing a gap between disks/heatmaps and distance-based signals. Moreover, geodesic-based signals have lower initial scores on both datasets and require more interactions.

\textbf{(M3) Consistent Improvement.} The consistent improvement is $\approx 65\%$ for both datasets, but it is slightly worse for AutoPET~\cite{autopet} as it is more challenging. Heatmaps and disks achieve the most consistent improvement, which means they are more precise in correcting errors. In contrast, geodesic distances change globally with new clicks as the whole guidance must be recomputed. These changes may confuse the model and lead to inconsistent improvement.

\textbf{(M4) Overlap with Ground Truth.} Heatmaps, disks, and EDT have a significantly higher overlap with the ground truth compared to geodesic-based signals, particularly on AutoPET~\cite{autopet}. GDT incorporates the changes in voxel intensity, which is not a strong signal for lesions with weak boundaries in AutoPET~\cite{autopet}, resulting in a smaller overlap with the ground truth. The guidances are ranked in the same order in \textbf{(M3)} and in \textbf{(M4)} for both datasets. Thus, a good overlap with the ground truth can be associated with precise corrections.

\textbf{(M5) Efficiency.} Efficiency is much higher on MSD Spleen~\cite{MSD-Spleen} compared to AutoPET~\cite{autopet}, as AutoPET has a $\times$2.4 larger mean volume size. The time also includes the sampling of new clicks for each simulated interaction. Disks are the most efficient signal, filling up spheres with constant values, while heatmaps are slightly slower due to applying a Gaussian filter over the disks. Distance transform-based guidances are the slowest on both datasets due to their complexity, but all guidance signals are computed in a reasonable time $(< 1s)$.

\textbf{Adaptive Heatmaps: Results.} Varying \textbf{(H1)-(H4)} and examining \textbf{(M1)-(M5)}, we find disks/heatmaps as the best signals, but with inflexibility near edges due to their fixed radius (Fig. \ref{fig:results}a)). Using GDT as a proxy signal to adapt $\sigma_i$ for each click $c_i$ mitigates this weakness by imposing large $\sigma_i$ in homogeneous areas and small, precise $\sigma_i$ near edges (Fig. \ref{fig:results}a)). This results in substantially higher consistent improvement and overlap with ground truth and the best initial and final Dice (Tab. \ref{tab:sigma_theta_results}). Thus, our comparative study has led to the creation of a more consistent and flexible signal with a slight performance boost, albeit with an efficiency cost due to the need to compute both GDT and heatmaps. 


\section{Conclusion}
Our comparative experiments yield insights into tuning existing guiding signals and designing new ones. We find that smaller radiuses ($\sigma \leq 5$), a small threshold ($\theta=10\%$), more iterations with interactions ($p \geq 75\%$), and traditional concatenation should be used. Weaknesses in existing signals include overly large radiuses near edges and inconsistent improvement for geodesic-based signals that change with each click. This analysis inspires our adaptive heatmaps, which adapt the radiuses of the heatmaps according to the geodesic values around the clicks, mitigating the inflexibility and inconsistency of existing guidances. We emphasize the importance of guidance representation in clinical applications, where a consistent and robust model is critical. Our study provides an overview of potential pitfalls, important parameters to tune, and how to design future guidance signals, along with proposed metrics for systematic comparison.

\section{Acknowledgements}
The present contribution is supported by the Helmholtz Association under the joint research school “HIDSS4Health – Helmholtz Information and Data Science School for Health. This work was performed on the HoreKa supercomputer funded by the Ministry of Science, Research and the Arts Baden-Württemberg and by the Federal Ministry of Education and Research.

\end{document}